\pdfoutput=1

\documentclass[letterpaper]{article}

\title{Synthesizing Action Sequences for Modifying Model Decisions}

\author{Goutham Ramakrishnan\thanks{Equal Contribution} , Yun Chan Lee\footnotemark[1] , Aws Albarghouthi\\ 
University of Wisconsin - Madison\\ 
\{gouthamr,aws\}@cs.wisc.edu, yunchan.c.lee@gmail.com
}

\usepackage{amsmath}
\usepackage{amsfonts}
\usepackage{amssymb}
\usepackage{dsfont}
\usepackage{mathtools}
\usepackage{xspace}
\usepackage{xcolor}
\usepackage{etoolbox}
\usepackage{pgfplots}
\usepgfplotslibrary{groupplots}
\usepackage{pgffor}
\usepackage{listings}
\usepackage{stmaryrd}
\usepackage{algorithm}  
\usepackage[noend]{algpseudocode}
\usepackage[inline]{enumitem}
\usepackage[square,numbers]{natbib}
\usepackage{booktabs}
\usepackage[group-separator={,}]{siunitx}
\usepackage{graphicx}
\usepackage{subcaption}
\usepackage{wrapfig}
\usepackage[font=small,labelfont=bf]{caption}

\newlist{mylist}{enumerate*}{1}
\setlist[mylist]{label=({\it\roman*})}

\renewcommand{\leq}{\leqslant}
\renewcommand{\geq}{\geqslant}

\DeclareMathOperator*{\argmin}{arg\,min}
\DeclareMathOperator*{\mean}{mean}

\renewcommand{\paragraph}[1]{\vspace{.1in}\noindent\textbf{{#1}.~~}}

\newcommand{\seq}{\sigma}
\newcommand\pair[1]{\langle#1 \rangle}
\newcommand{\actions}{A}
\newcommand{\params}{R}
\newcommand{\param}{r}
\newcommand{\pseq}{\rho}

\newcommand{\action}{a}
\newcommand{\cost}{c}
\newcommand{\pre}{\mathit{pre}}

\newcommand{\model}{f}
\newcommand{\dom}{X}
\newcommand{\inst}{x}

\begin{document}

\maketitle

\begin{abstract}
When a model makes a consequential decision, e.g., denying someone a loan, it needs to additionally generate actionable, realistic feedback on what the person can do to favorably change the decision. We cast this problem through the lens of program synthesis, in which our goal is to synthesize an optimal (realistically cheapest or simplest) sequence of actions that if a person executes successfully can change their classification. We present a novel and general approach that combines search-based program synthesis and test-time adversarial attacks to construct action sequences over a domain-specific set of actions. We demonstrate the effectiveness of our approach on a number of deep neural networks.
\end{abstract}

\section{Introduction}\label{sec:introduction}

Today, predictive models are responsible for an ever-expanding spectrum of decisions, some of which are consequential to the lives and well-being of individuals---e.g., mortgage underwriting, job screening, healthcare decisions, criminal risk assessment, and many more. 
As such, questions about fairness and transparency have taken center stage in the debate over the increasing use of machine learning to automate decisions in sensitive domains, and a vibrant research community has emerged to explore and address the many facets of these questions.

In this paper, we are interested in the problem of providing actionable feedback to the subjects of algorithmic decision-making. For instance, imagine that you are denied a mortgage to buy your first home, thanks to a model that consumed a set of features and deemed you too risky. 
We envision that such an algorithmic process should additionally give you realistic, actionable feedback that will increase your chances of receiving a loan. For instance, you might receive the following feedback: \emph{increase your down payment by \$1000} and \emph{limit credit card debt to a maximum of \$5000 for the next two months}. 
This is probably reasonable advice, in contrast with, say, a much harder to fulfill feedback like \emph{change your marital status from single to married}.

We view this problem through the lens of \emph{program synthesis}~\cite{PGL-010}: we want to synthesize an \emph{optimal sequence of instructions} that a human can \emph{execute} so that they favorably change the decision of some model. Optimality here is with respect to a measure of how hard it is for a person to perform the provided actions---we want to provide the simplest, cheapest feedback.
There are many challenges in solving this problem: 
\begin{mylist}
    \item the combinatorial blow-up in the space of action sequences,
    \item the fact that actions are parameterized by real values and have variable cost (e.g., \emph{increase savings by \$X}),
    \item and the fact that action ordering is important, e.g., \emph{you can only do A after you have done B}
    or \emph{you can only do A if you are more than 35 years old}.
\end{mylist}

To attack this problem, we make the key observation that the problem resembles that of generating \emph{adversarial examples}~\cite{szegedy2013intriguing,GoodfellowSS14,carlini2017towards,papernot2017practical}, where we usually want to slightly perturb the pixels of an input image to modify the classification, e.g., make a neural network think a dog is a panda. However, in our case, we are \emph{not} looking for an imperceptible perturbation to the input features,
but one that results from the application of real-world actions.
With this view in mind, we present a new technique that adapts and combines \begin{mylist} \item \emph{search-based program synthesis}~\cite{alur2018search} to traverse the space of action sequences and \item \emph{optimization-based adversarial example generation}~\cite{carlini2017towards} techniques to discover action parameters. \end{mylist}
This combination allows our approach to handle a rich class of (differentiable) models, e.g., deep neural networks, and complex domain-specific actions and cost models.

\paragraph{Setting and Consequences}
At this point, it is important to recognize the possibility that the solution we propose for the problem setting may be vulnerable to unethical practices. 
Although our technique, in principle, may be used by users to maliciously \emph{game} the system, we believe in its importance and cannot envision a world in which subjects are unable to understand and \textit{act} on black-box decisions. 
One setting we envision is where users cannot adversarially attack the decision-maker as they do not have access to the model. The intention would be to use our technique as a means for the service provider to give meaningful and actionable feedback to its users, making the decision process more transparent. 

\paragraph{Most Relevant Work}
To our knowledge, the idea of providing actionable feedback for algorithmic decisions was first advocated by Wachter et al.~\cite{wachter2017counterfactual} in a law article.
 Ustun et al.~\cite{ustun2019actionable} implemented this idea by searching for a minimal change to input features to modify the classification of simple linear models (logistic regression).
 Their approach discretizes the feature space and encodes the search as an integer programming (IP) problem.
 Zhang et al.~\cite{zhang2018interpreting} consider a similar problem over neural networks composed of ReLUs, exploiting the linear structure of ReLUs to solve a series of LP problems to construct a convex region of positively classified points that are close to the input.
Our work is different in a number of dimensions: 
\begin{mylist}
    \item Our algorithm is quite general: by reducing the search to an optimization problem, \`a la test-time adversarial attacks, it can handle the general class of differentiable models (as well as differentiable action and cost definitions), instead of just linear models. 
\item We allow defining complex, nuanced actions that mimic real-world interventions, as opposed to arbitrary modifications to input features. 
\item Similarly, we allow encoding cost models to assign different costs to actions, with the goal of synthesizing the simplest feedback a person can act on.
\end{mylist}
See Section~\ref{sec:relatedwork} for an extended discussion of related work. 

\paragraph{Contributions}
Our contributions are:

\begin{itemize}
    \item We define the problem of \emph{synthesizing optimal action sequences} that can favorably change the output of a machine-learned model. We view the problem through the lens of program synthesis, where our goal is to synthesize a program over a domain-specific language of real-world actions specified by a domain expert. 
    \item We present an algorithm that combines \emph{search-based program synthesis} to traverse the space of action sequences and \emph{optimization-based test-time adversarial attacks} to discover optimal parameters for action sequences.
    We demonstrate how to leverage the results of optimization to guide the combinatorial search.
    \item We implement our approach and apply it to a number of neural networks learned from popular datasets.
    Our results demonstrate the effectiveness of our approach, the benefits of our algorithmic decisions,
    and the robustness of the synthesized action sequences to noise.
\end{itemize}
\section{Optimal Action Sequences}\label{sec:problem}

In this section, we formally define the problem of synthesizing optimal action sequences. 

\paragraph{Decision-Making Model}
We shall use $\model: \dom \rightarrow \{0,1\}$ to denote a classifier over inputs in $\dom$.
For simplicity of exposition, and without loss of generality, we restrict $\model$ to be a binary classifier---%
our approach extends naturally to $k$-ary classifiers.

\paragraph{A DSL of Actions}
For a given classification domain, we assume that we have a \emph{domain-specific} set of actions $\actions = \{a_1,\ldots,a_n\}$, perhaps curated by a domain expert.
Each action $\action \in \actions$ is a function $\action: \dom \times \params \rightarrow \dom$,
where $\params$ is the set of parameters that $\action$ can take.
For example, imagine $\inst \in \dom$ are features of a person applying for a loan.
An action $\action(\inst,1000)$ may be one that increases $\inst$'s savings by $\$1000$,
resulting in $\inst' \in \dom$.

For each action $\action_i \in \actions$, we associate a cost function $\cost_i: \dom \times \params \rightarrow \mathds{R}_{\geq 0}$, denoting the cost of applying $\action_i$ on a given input and parameters.
Making $\cost_i$ a function of inputs and parameters of an action allows us to define fine-grained cost functions, e.g., some actions may be easier for some people, but not for others. For instance, in the US, acquiring a credit card is much easier for someone with a credit history in contrast to someone who recently arrived on a work visa.
Similarly, varying the parameter of an action should vary the cost, e.g., 
\emph{increase savings by \$1000} should be much cheaper than \emph{increase savings by \$1,000,000}.

Additionally, for each action $\action_i$, we associate a Boolean \emph{precondition} $\pre_i: \dom \times \params \rightarrow \mathds{B}$, indicating whether action $\action_i(\inst,\param)$ is \emph{feasible} for a given input $\inst$ and parameter $\param$.
There are a number of potential use cases for preconditions.
For instance, the action of renting a car may be only allowed if you are over 21 years old; this can be encoded as the  precondition $\emph{age} \geq 21$. 
Preconditions can also encode valid parameters, e.g., you cannot increase your credit score past 850, so an action which recommends increasing your credit score by $\param$ will have the precondition $\emph{creditScore} + \param \leq 850$.

\paragraph{Optimal Action Sequence}
Fix an input $\inst\in\dom$ and assume that $\model(\inst) = 0$.
Informally, our goal is to find the least-cost, feasible sequence of actions that can transform $\inst$ into an $\inst'$
such that $\model(\inst') = 1$.

Formally, we will define an action sequence using a pair of sequences $\pair{\seq,\pseq}$, denoting actions in $\actions$ and their corresponding parameters in $\params$, respectively.
Specifically,
$\seq$  is a sequence of integers in $[1,|\actions|]$ (action indices),
and $\seq_i$  denotes the $i$th element in this sequence.
 $\pseq$ is a sequence of parameters such that each $\pseq_i \in \params$.
We assume $|\pseq| = |\seq| = k$, and we will use $k$ throughout to denote $|\seq|$.

Given pair $\pair{\seq,\pseq}$, in what follows, we will use
 $\inst_i = a_{\seq_i}(\inst_{i-1},\pseq_i)$, where $i \in [1,k]$ and $\inst_0 = x$. That is, 
variable $\inst_i$ refers to the result of applying the first $i$ actions defined by $\pair{\seq,\pseq}$ to the input $\inst$.
We are therefore looking for a feasible, least-cost sequence of actions $\action_{\seq_1},\ldots,\action_{\seq_k}$
and associated parameters $\pseq_1,\ldots,\pseq_{k}$,
which, if applied starting at $\inst$, results in $\inst_{k}$ that is classified as 1 by $\model$.
This is captured by the following optimization problem:\footnote{Equivalently, we can cast this as an optimal \emph{planning} problem, where $\inst$ is the initial state, our goal state $\inst_k$ is one where $f(\inst_k)=1$, and actions transition us from one state to another.}
%
\begin{align}
    \argmin_{\pair{\seq,\pseq}} &  \sum_{i=1}^{k} c_{\seq_i}(\inst_{i-1},\pseq_i) \label{eq:main}\\
    \text{subject to } & f(\inst_{k}) = 1 \text{ and } 
                \bigwedge_{i=1}^{k} \pre_{\seq_i}(\inst_{i-1},\pseq_i) \nonumber
\end{align}

\section{An Algorithm for Sequence Synthesis}\label{sec:algorithm}

\newcommand{\smodel}{g}
\newcommand{\objective}{d}
\newcommand{\score}{\mathit{score}}
\newcommand{\vscore}{\score_v}
\newcommand{\gscore}{\score_g}
\newcommand{\oscore}{\score_o}

\newcommand{\fp}{\emph{fp}}

\newcommand{\seqset}{S}

\begin{algorithm*}[t!]
  \caption{Full synthesis algorithm}\label{alg:search}
  \begin{algorithmic}[1]
    \Function{synthesize}{model $\model$, instance  $\inst$, actions $\actions$}
      \State $\seqset \gets \{\pair{\seq^\emptyset, \pseq^\emptyset}\}$, where $\pair{\seq^\emptyset,\pseq^\emptyset}$ are the empty action and parameters sequences, respectively
      \Repeat
          \State Let $\pair{\seq,\pseq} \in \seqset$ and $\action_i \in \actions$ be with smallest $\score(\seq,\pseq,\action_i)$ \Comment $\pair{\seq\action_i,-} \not\in \seqset$
          \State Solve Problem~\ref{eq:cw} to compute parameters $\pseq'$ for sequence $\seq\action_i$
          \State $\seqset \gets \seqset \cup \{\pair{\seq\action_i,\pseq'}\}$
      \Until {threshold exceeded} \Comment \emph{threshold can be, e.g., search depth}
      \State \Return Solution of Problem~\ref{eq:main} restricted to sequences in $\seqset$
    \EndFunction
  \end{algorithmic}
\end{algorithm*}

We now present our technique for synthesizing action sequences, based on the optimization objective outlined above.  
Our algorithm assumes a differentiable model, e.g., a deep neural network,
of the form $\model: \mathds{R}^m \rightarrow \{0,1\}$,
as well as differentiable actions, cost functions and preconditions. 
To solve Problem \ref{eq:main},  defined in Section~\ref{sec:problem},
we break it into two pieces:
\begin{mylist}\item a discrete search through the space of action sequences $\seq$ and \item a continuous-optimization-based search through the space of action parameters $\pseq$, which we assume to be real-valued.\end{mylist}

In Section~\ref{ssec:opt}, we begin by describing the optimization technique, by assuming we have a fixed sequence of actions and setting up an optimization problem---an adaptation of Carlini and Wagner's adversarial attack~\cite{carlini2017towards}---to learn the parameters to those actions.
Then, in Section~\ref{ssec:synth}, we present the full algorithm, a search-based synthesis algorithm for discovering action sequences, which uses the optimization technique from Section~\ref{ssec:opt} as a subroutine.

\emph{Remark on optimality:}
We note that the constrained optimization Problem~\ref{eq:main} is hard in general---e.g., even very limited numerical planning problems that can be posed as Problem~\ref{eq:main} are undecidable~\cite{helmert2002decidability}. Our use of adversarial attacks in the following section  necessarily relaxes some of the constraints and is therefore not guaranteed to result in optimal action sequences.

\subsection{Adversarial Parameter Learning}
\label{ssec:opt}
We now assume that we have a fixed sequence of actions $\seq$,
as defined in Section~\ref{sec:problem}.
Our goal is to find the parameters $\pseq$ such that $\pair{\seq,\pseq}$ satisfies the constraints of Problem~\ref{eq:main}.
Specifically, our solution to this problem is an adaptation of Carlini and Wagner's seminal adversarial attack technique against neural networks~\cite{carlini2017towards}, but in a setting where the ``attack'' is comprised of a sequence of actions with preconditions and varying costs.

Henceforth, we shall assume that the model is a neural network $\model: \mathds{R}^m \to \mathds{R}^2$
(where the output denotes a distribution over the two classification labels).
Additionally, $\model(x) = \text{softmax}(\smodel(x))$, i.e., function $\smodel$ is the output of the pre-softmax layers of the network.

\paragraph{Boolean Precondition Relaxation}
Our goal is to construct a tractable optimization problem whose solution results in the parameters $\pseq$
to the given action sequence $\seq$.
We begin by defining the following constrained optimization problem which relaxes the Boolean precondition constraints:
\begin{align}
    \argmin_{\pseq} &~  \sum_{i=1}^{k} \cost_{\seq_i}(\inst_{i-1},\pseq_i) + \pre_{\seq_i}'(\inst_{i-1},\pseq_i) \label{eq:constrained}\\
     \text{subject to} &~ \model(x_{k})_1 > \model(x_{k})_0 \nonumber
\end{align}
where $\model(x)_j$ is the probability of class $j$,
and the function $\pre_i'$ is a continuous relaxation of the Boolean precondition $\pre_i$.  Specifically, we encode preconditions by imposing a high cost on violating them.
For instance, if $\pre_i(\inst,\param) = x > c$, where $c$ is a constant, then
we define $\pre'_i(\inst,\param) = \tau\exp(-\tau'(\inst-c))$, where $\tau$ and $\tau'$ are hyperparameters.\footnote{We assume that expressions in preconditions are differentiable.}
The hyperparameters $\tau$ and $\tau'$ determine the steepness of the continuous boundaries;
the values we choose are inversely proportional to the size of the domain of $\inst$.
We detail our specific choices in the Appendix.
Conjunctions of Boolean predicates are encoded as a summation of their relaxations.
We can now define $C_{\seq_i}(\inst_{i-1},\pseq_i)$ to be the overall cost incurred by the action-parameter pair $\pair{\seq_i,\pseq_i}$, i.e.
\begin{align}
    C_{\seq_i}(\inst_{i-1},\pseq_i) = \cost_{\seq_i}(\inst_{i-1},\pseq_i) + \pre_{\seq_i}'(\inst_{i-1},\pseq_i) \nonumber
\end{align}


\paragraph{Carlini--Wagner Relaxation}
Now that we have relaxed preconditions, what is left is the classification constraint $\model(x_{k})_1 > \model(x_{k})_0$ in Problem~\ref{eq:constrained}.
Following Carlini and Wagner, we transform the constraint $\model(x_{k})_1 > \model(x_{k})_0$ 
into the objective function that is the distance between logit (pre-softmax) output:
$ h(x_k) = max(0, g(x_{k})_0 - g(x_{k})_1)$.\footnote{
    Note that there many alternative relaxations of $\model(x_{k})_1 > \model(x_{k})_0$; Carlini and Wagner explore a number of alternatives, e.g.,  using $f$ instead of $g$, and show that this outperforms them.
}

This results in the following optimization problem:
\begin{align}
    \argmin_{\pseq} c \cdot h(x_k) + \sum_{i=1}^{k} C_{\seq_i}(\inst_{i-1},\pseq_i) \label{eq:cw}
\end{align}

In practice, we perform an adaptive search for the best value of the hyperparameter $c$ as we solve the optimization problem: at a variable length interval $t$ of minimization steps, we determine how close the search is to the decision boundary and adjust $c$ and $t$ accordingly. The Appendix details the exact algorithm we use for updating $c$ and $t$.

\begin{table*}[!t]
    \centering
    \small
    \begin{tabular}{ l  p{4cm}  c c  c}
        \toprule
        Dataset/model & Network architecture  & \#Features & \#Actions \\
        \midrule
        \emph{German} Credit Data & 2 dense layers of 40 ReLUs  & 20 & 7 \\
        \emph{Adult} Dataset & 2 dense layers of 50 ReLUs  & 14 & 6 \\
        \emph{Fannie Mae} Loan Perf. & 5 dense layers of 200 ReLUs  & 21 & 5 \\
        \hline
        \emph{Drawing} Recognition & 3 1D conv. layers, 1 dense layer of 1024 ReLUs & 512 (pixels)
        & 1\\
        \bottomrule
    \end{tabular}
    \caption{Overview of datasets/models for evaluation; we will refer to a model by the italicized prefix of its name}
    \label{tab:datasets}
\end{table*}

\subsection{Sequence Synthesis and Optimization}
\label{ssec:synth}
Now that we have defined the optimization problem for discovering parameters $\pseq$ of a given action sequence $\seq$,
we proceed to describe the full algorithm, where we search the space of action sequences.

\paragraph{Algorithm Description}
Algorithm~\ref{alg:search} is a simple search guided by a parametric \emph{score} function that directs the search---lower score is better.
 The algorithm maintains a set of sequences $\seqset$, which initially is the pair $\pair{\seq^\emptyset, \pseq^\emptyset}$, containing two empty sequences.
 In every iteration, the algorithm picks an action sequence in $\seqset$, extends it with a new action from $\actions$,
 and solves  optimization Problem~\ref{eq:cw} to compute a new set of parameters.
 The search process continues until some preset threshold is exceeded, e.g., we have covered all sequences of some length or we have discovered a sequence that is below some cost upper bound.
Finally, we can return the best pair in $\seqset$, i.e., the one with the minimal cost that changes the classification and satisfies all preconditions, as per Problem~\ref{eq:main}.

  \paragraph{Defining the Scoring Function}
  The definition of the scoring function $\score$ dictates the speed with which
  the algorithm arrives at an best action sequence.
  In our evaluation, we consider a number of definitions, the first of which, the \emph{vanilla} definition, is simple, but often inefficient: $$\vscore(\seq,\pseq,\action_i) = k + 1$$ This definition turns the search into a breadth-first search, as shorter sequences are evaluated first.

  A more informed score function we consider is to simply return the value of the objective function in Problem~\ref{eq:cw}
  for a given sequence $\pair{\seq,\pseq}$. We call this function $\oscore$. Notice that $\oscore$ does not consider the action to apply, so the action with which to expand the sequence is chosen arbitrarily.

  Next, we consider a more sophisticated scoring function:
  we want to pick the action that modifies the most important features.
  To do so, we use the gradient of model features, with respect to the target loss, as a proxy for the most important features.
  The idea is that we want to pick the action that modifies the features with the largest gradient.
  For every action $\action_i \in \actions$, we use the $\fp(\action_i)$
  to denote its \emph{footprint}: the set of indices of the input features it can modify, e.g., $\fp(\action_i) = \{1,2\}$
  means that it modifies features 1 and 2 and leaves all others unchanged.
  Given $\pair{\seq,\pseq} \in \seqset$, let $\inst'$ be the result of
  applying $\pair{\seq,\pseq}$ to the input instance $\inst$.
  We now define the score function as:
  $$\gscore(\seq,\pseq,\action_i) = -\mean_{j \in \fp(\action_i)}  \left| \frac{d\ell(\model(\inst'))}{dx_j} \right|$$
  In other words, the score of applying $\action_i$ after the sequence $\seq$ depends on 
  the average gradient of the target loss $\ell$---binary cross-entropy loss with respect to the target label, i.e., 1---with respect to the features in $\action_i$'s footprint.

\section{Implementation and Evaluation}\label{sec:evaluation}

\newcommand{\qu}[1]{\emph{\textbf{Q#1}}}

\paragraph{Implementation}
Our algorithm is implemented in Python 3, using TensorFlow~\cite{abadi2016tensorflow}. 
Actions, along with their costs and preconditions, are implemented as instances of an \texttt{Action} Python class.
The Adam Optimizer~\cite{kingma2014adam} is used to solve optimization Problem~\ref{eq:cw}.
For fast experimentation, we implemented a brute-force version of Algorithm~\ref{alg:search}
where all sequences up to some length $n$ are optimized in parallel using AWS Lambda---i.e., each sequence is optimized as a separate Lambda.

\paragraph{Research Questions}
We have designed our experiments to answer the following research questions:
\qu{1}: Can our technique synthesize action sequences for non-trivial models and actions?
\qu{2}: How do different score functions impact algorithm performance?
\qu{3}: How robust are the synthesized action sequences to noise?
Further, \qu{4}: we explore other applications of our technique, beyond consequential decisions.

\paragraph{Domains for Evaluation}
For exploring questions \qu{1-3}, we consider three popular datasets: The German Credit Data~\cite{Dua:2019} and the Fannie Mae Single Family Loan Performance~\cite{mae2014fannie} datasets have to do with evaluating loan applications---high or low risk. 
The Adult Dataset~\cite{Dua:2019} predicts income as high or low---the envisioned use case is it can be used to set salaries.
Table~\ref{tab:datasets} summarizes our datasets and models: For each of the three datasets, we 
\begin{mylist}
    \item trained a deep neural network for classification (in the case of the Fannie Mae dataset, we used the neural network from~\cite{zhang2018interpreting}),
    \item constructed a number of realistic actions along with their associated costs and preconditions,
    and \item randomly chose 100 negatively classified instances (i.e., 300 instances overall) from the test sets to apply our algorithm to.
\end{mylist}

The actions constructed for each domain cover both numerical and categorical features;
a number of actions for each domain modify multiple features---e.g., change the debt-to-income ratio,
or get a degree (which takes time and therefore increases age).
The Appendix details all the actions and describes our encoding of actions that modify categorical features.

To explore \qu{4}, we also consider the Drawing Recognition task~\cite{zhang2018interpreting} based on Google's \emph{Quick, Draw!} dataset~\cite{quickdraw}.
The goal is to extend a drawing, represented as a set of straight lines, so as it is classified as, e.g., a cat.
Hence, we only build one action for this model: add line from point $(a,b)$ to $(a',b')$.

\subsection{Results}\label{ssec:results}
We are now ready to discuss the results. Henceforth, when we refer to \emph{optimal solution},
we mean the best solution we find after Algorithm~\ref{alg:search} has explored all sequences
of length less than an upper bound.

\begin{figure}[t]
    \centering
    \begin{subfigure}{.5\textwidth} 
        \centering
        \includegraphics[scale=0.18]{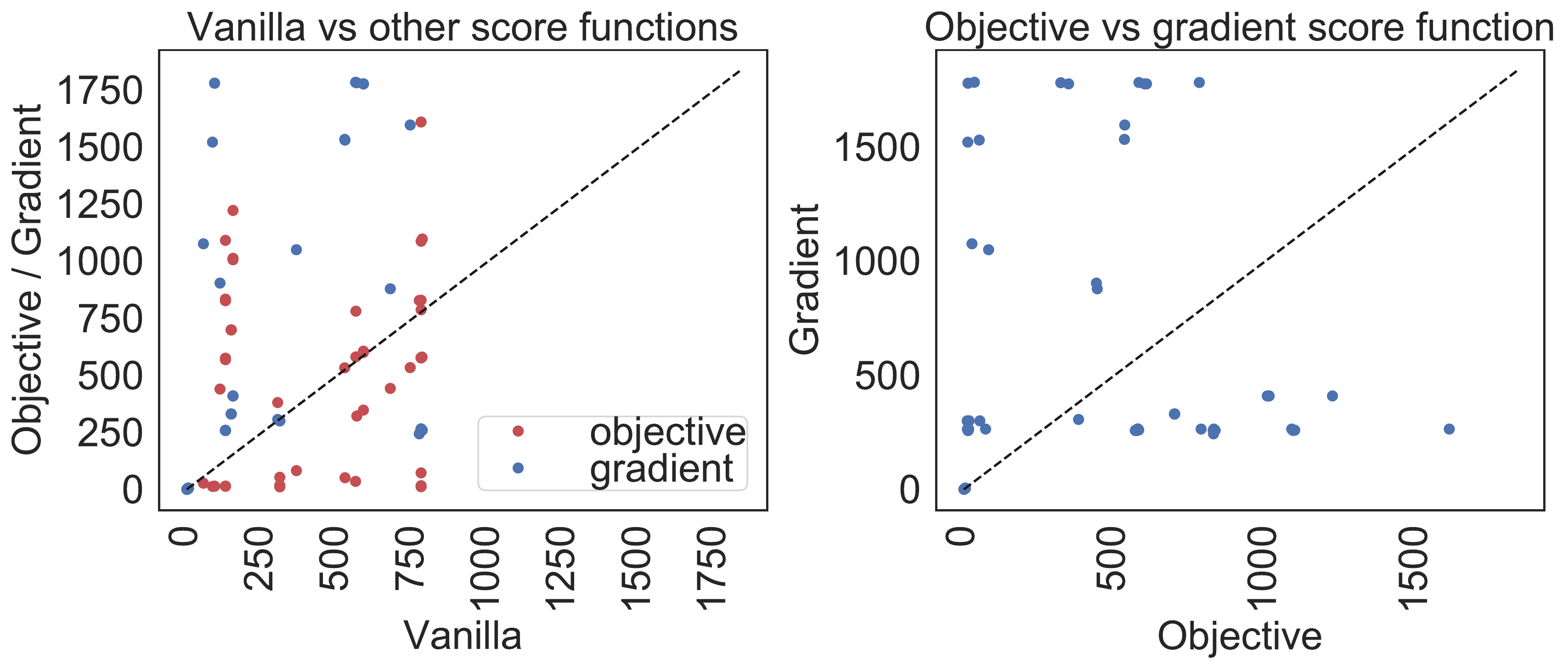}
        \caption{Results for the German model}
    \end{subfigure}
    \begin{subfigure}{0.49\textwidth} 
        \centering
        \includegraphics[scale=0.18]{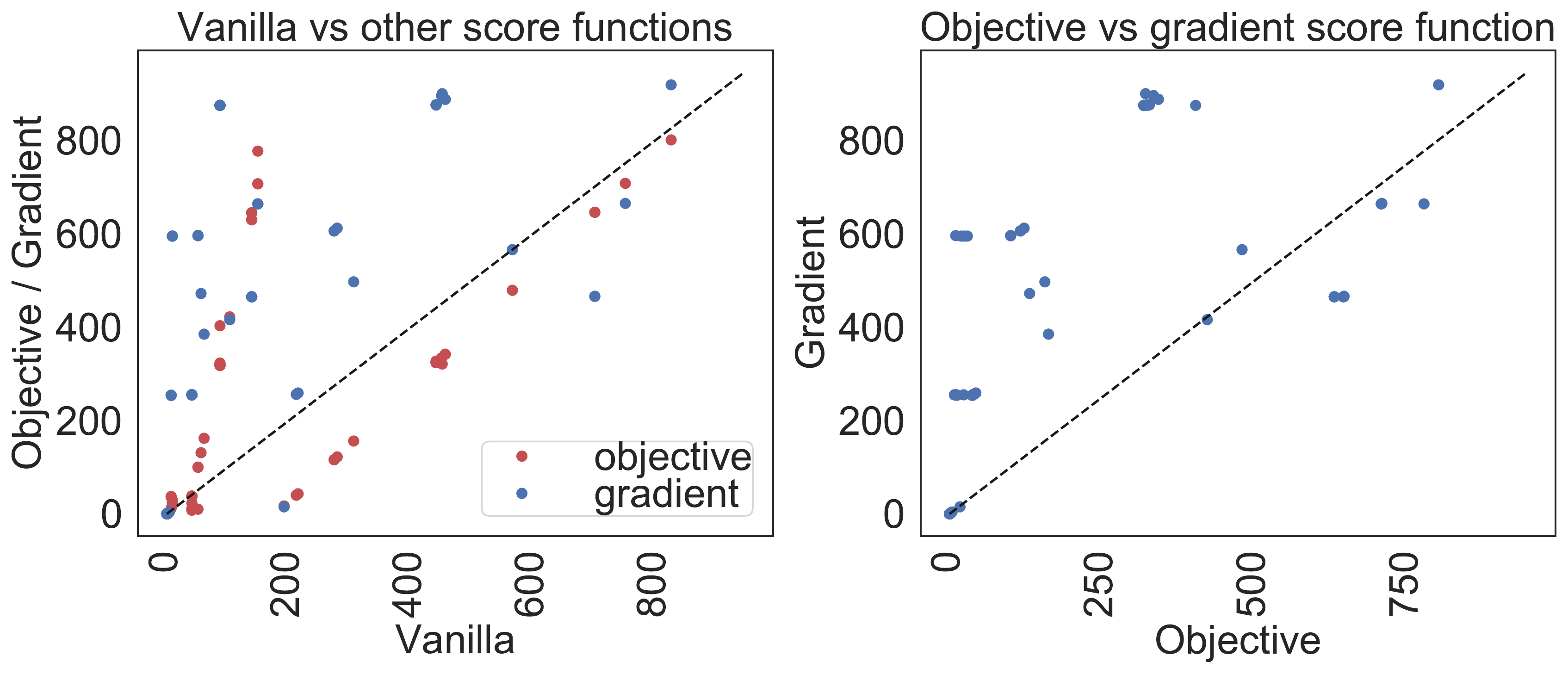}
        \caption{Results for the Adult model}
    \end{subfigure}

    \begin{subfigure}{1\textwidth}
        \centering
        \includegraphics[scale=0.18]{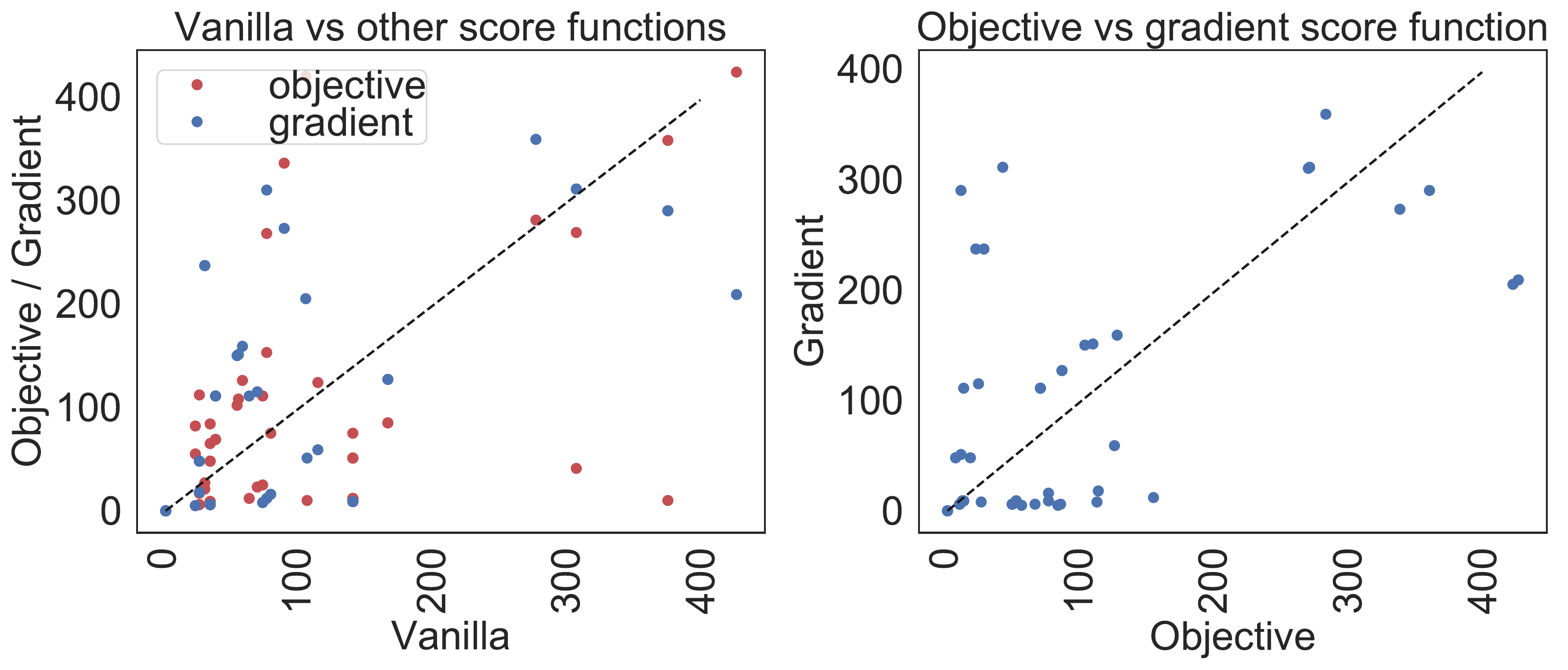}
        \caption{Results for the Fannie Mae model}
    \end{subfigure}
    \caption{Number of optimization problems solved---i.e., loop iterations of Algorithm~\ref{alg:search}---before arriving at the optimal solution for different score functions.
    Each dot represents an instance.}\label{fig:main}
\end{figure}

\paragraph{Instances Solved}
For our primary models, we make our algorithm consider all sequences of length $\leq 4$.
The rationale behind this choice is that we usually want a small set of instructions to provide to an individual.
Our algorithm was able to find solutions to 100/100 instances in the German model, 90/100 instances in the Adult model,
and 62/100 instances in the Fannie Mae model.
Note that inability to find a solution could be due to insufficient actions or incompleteness of the search---%
sequence length limit, relaxation of optimization problem, or local minima. In particular, the relatively inferior success rate on the Fannie Mae model may be a direct result of the fact that the neural network is much deeper. 

To give an example of a synthesized action sequence by our algorithm, consider the following
sequence of 3 actions for Fannie Mae:
\emph{Increase credit score by 17 points}, \emph{Reduce loan term by 43 months}, 
and \emph{Increase interest rate by 0.621}.

\begin{wrapfigure}{r}{0.4\textwidth}
    \vspace{-.3in}
    \begin{center}
      \includegraphics[width=0.33\textwidth]{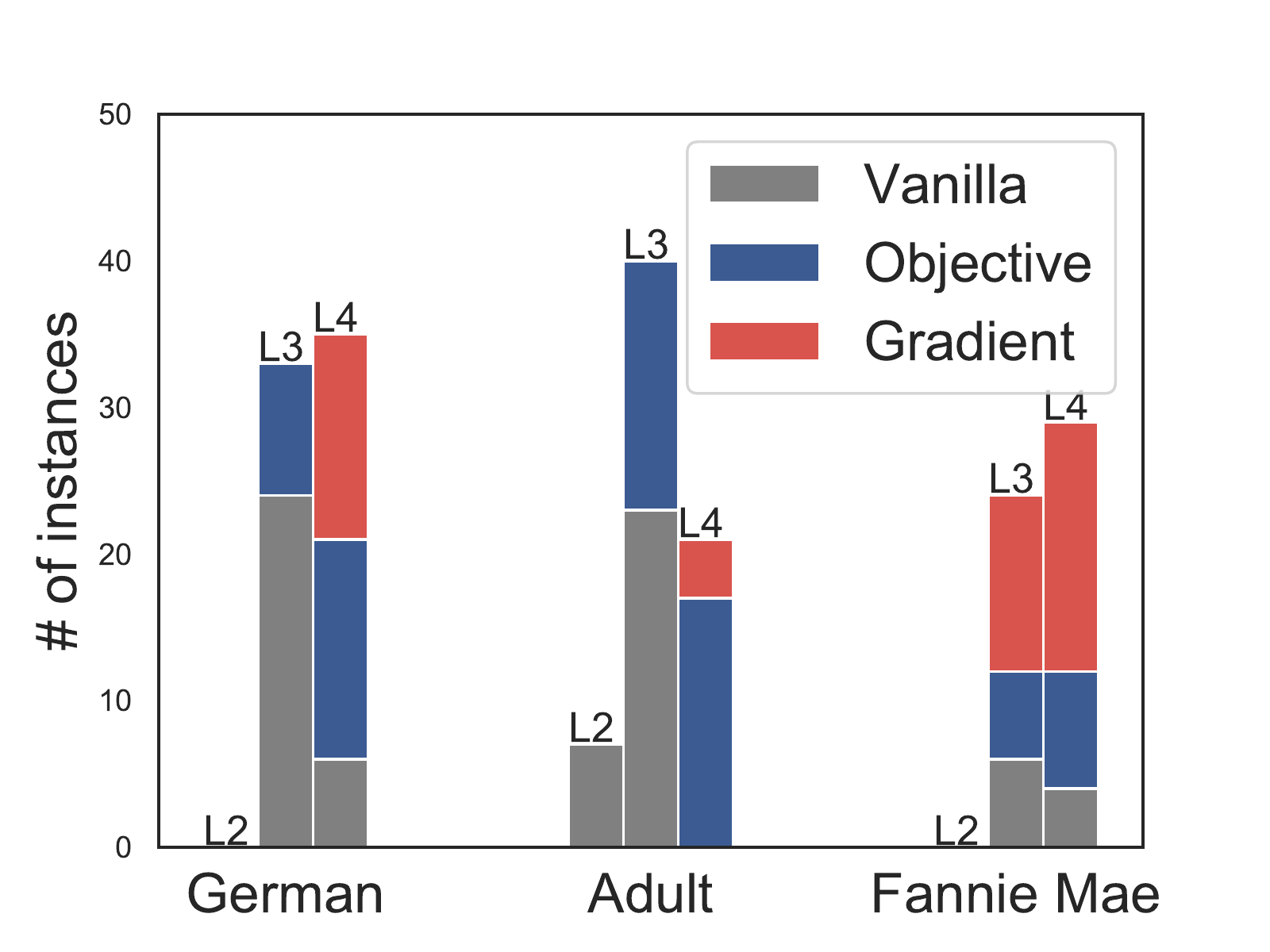}
    \end{center}
    \vspace{-.1in}
    \caption{Performance of score functions as length of optimal sequence increases---L$x$ denotes optimal sequence of length $x$}\label{fig:length}
    \vspace{-.1in}
  \end{wrapfigure}
  
\paragraph{Effects of Score Function}
Next, we explore the effects of different score functions.
Recall, in Section~\ref{ssec:synth}, we defined the \emph{vanilla} score function $\vscore$, 
where sequences are explored by length (a breadth-first search);
the \emph{objective} score function $\oscore$, where the sequence with the smallest solution to Problem~\ref{eq:cw} is explored; and the \emph{gradient} score function $\gscore$, where gradient of cross-entropy loss is used to choose the sequence and action to explore.

Figure~\ref{fig:main} shows the results of the German, Adult, and Fannie Mae models.
Each point is one of the instances and the axes represent the iteration at which
a score function arrived at the optimal sequence.\footnote{Time/iteration is $\sim$15s across instances; we thus focus on number of iterations as performance measure.}
The left plot compares $\vscore$ vs $\gscore$ (blue) and $\oscore$ (red);
the right plot compares $\oscore$ vs $\gscore$.
We make two important observations:
First, the vanilla score function excels on many instances (points above diagonal).
After investigating those instances, we observe that they have short optimal sequences---of length 1 or 2.
This is perhaps expected, as $\oscore$ and $\gscore$ may quickly lead the search towards longer sequences,
missing short optimal ones until much later.
We further illustrate this observation in Figure~\ref{fig:length},
where we plot the number of times each score function outperformed others (in terms of number of iterations of Algorithm~\ref{alg:search}) when the optimal solution
is of length 2,3, and 4.
We see that when optimal sequences are longer, $\vscore$ stops being effective.
Second, we observe that both the gradient and objective score functions, $\gscore$ and $\oscore$, have their merits, e.g., for Fannie Mae, $\gscore$ dominates while for Adult $\oscore$ dominates.

\begin{figure*}[t]
    \centering
    
    \begin{subfigure}{0.3\textwidth} 
        \centering
        \includegraphics[scale=0.26]{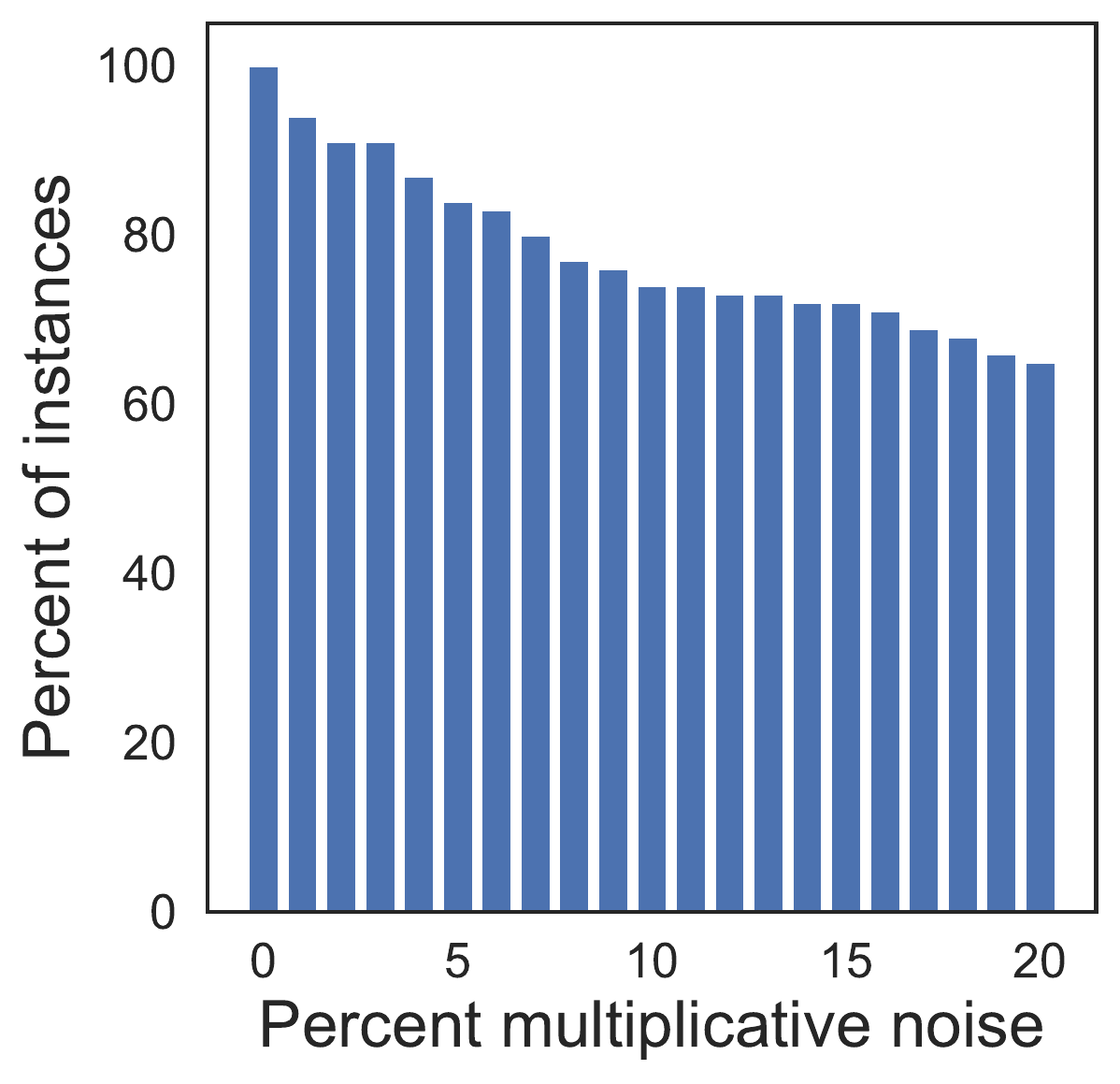}
        \caption{German}
    \end{subfigure}
    \begin{subfigure}{0.3\textwidth} 
        \centering
        \includegraphics[scale=0.26]{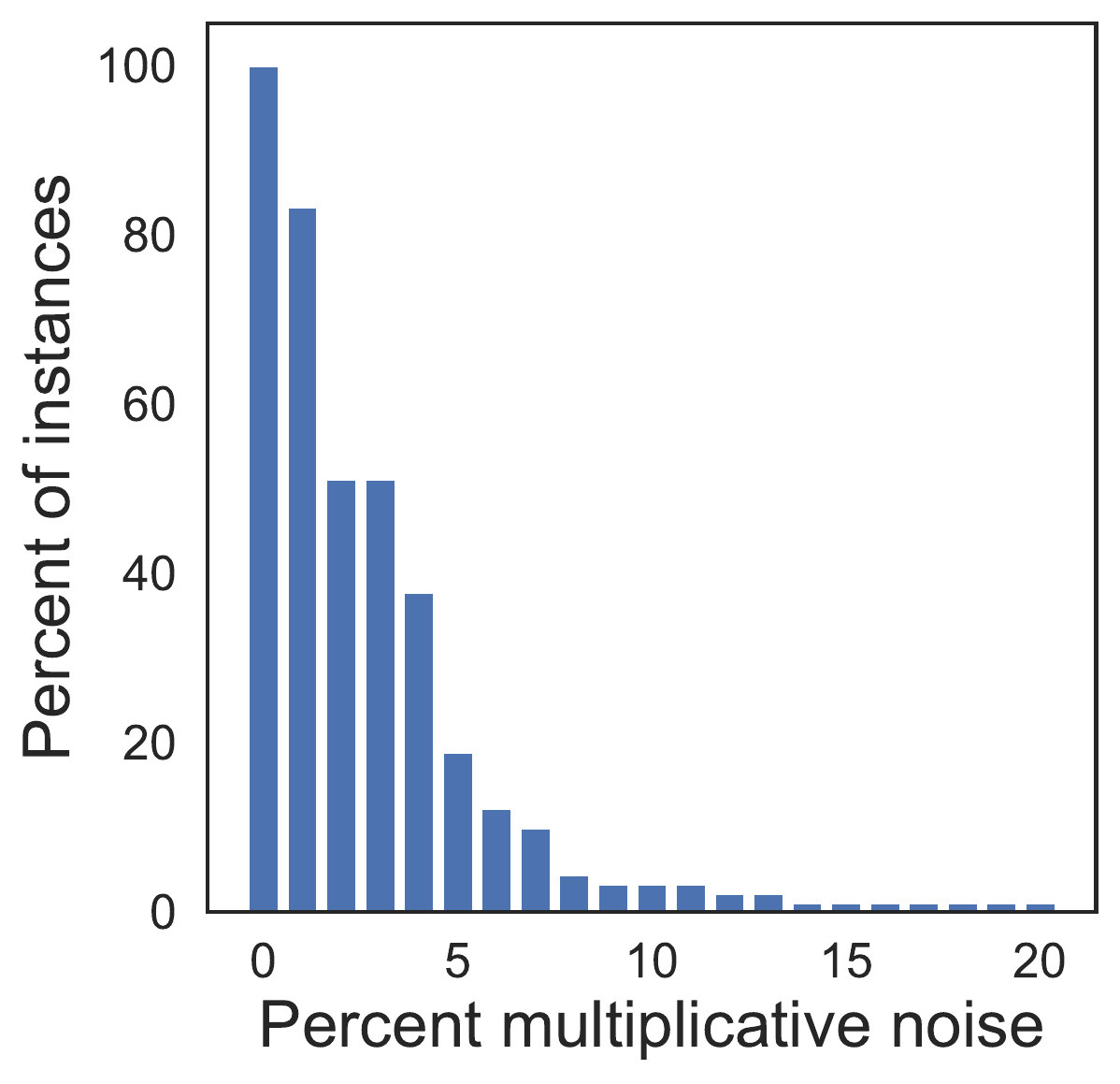}
        \caption{Adult}
    \end{subfigure}
    \begin{subfigure}{0.3\textwidth} 
        \centering
        \includegraphics[scale=0.26]{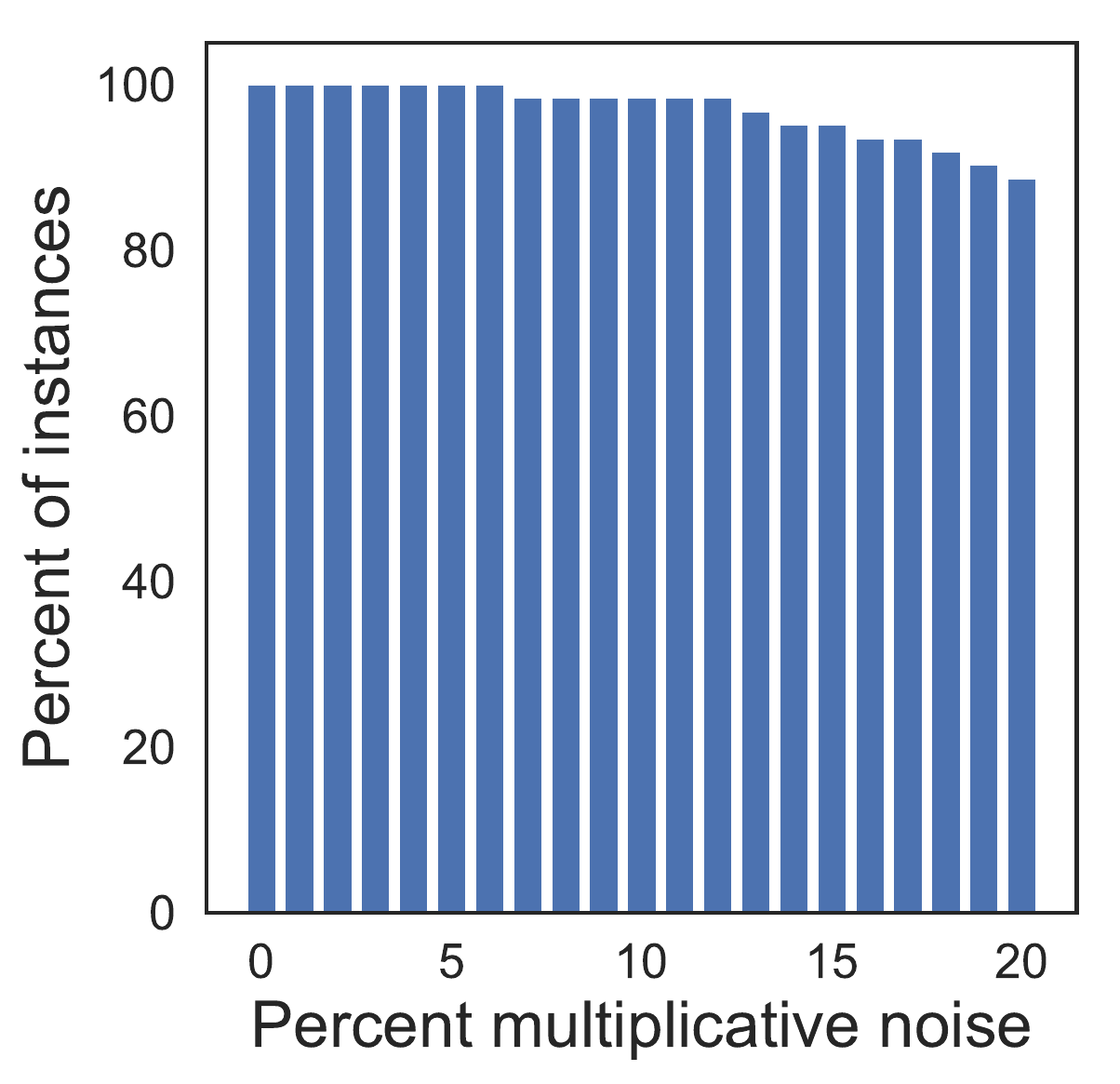}
        \caption{Fannie Mae}
    \end{subfigure}
    \caption{Percent of instances ($y$-axis) that tolerate noise $\theta$ ($x$-axis) noise with probability $\geq 0.8$}\label{fig:robust}
\end{figure*}

\begin{figure*}[t]
    \centering
    
    \begin{subfigure}{0.32\textwidth} 
        \centering
        \includegraphics[scale=0.24]{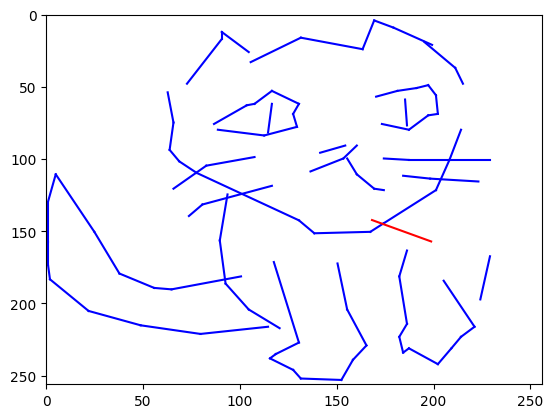}
        \caption{sequence length = 1}
    \end{subfigure}
    \begin{subfigure}{0.32\textwidth} 
        \centering
        \includegraphics[scale=0.24]{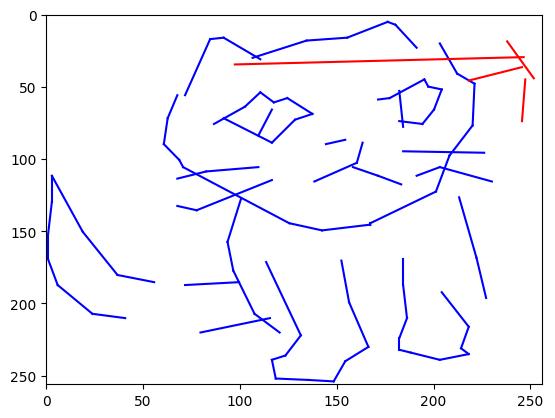}
        \caption{sequence length = 4}
    \end{subfigure}
    \begin{subfigure}{0.32\textwidth} 
        \centering
        \includegraphics[scale=0.24]{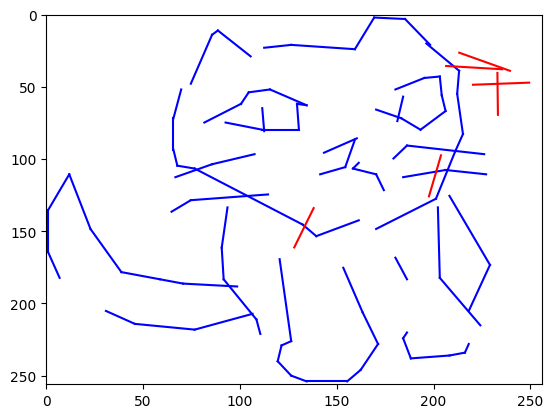}
        \caption{sequence length = 6}
    \end{subfigure}
    \caption{Three \emph{different} Quick, Draw! examples. Blue strokes comprise the input instance; red strokes are the results of applying actions.}\label{fig:sketch}
\end{figure*}

\paragraph{Robustness of Synthesized Sequences}
We now investigate how robust our solutions are to perturbations
in their parameters.
The idea is that a person given feedback from our algorithm may not be able to 
fulfil it precisely.
We simulate this scenario by adding noise to the parameters of the optimal sequences.
Specifically, for each synthesized optimal sequence and each parameter $\param$ in the sequence,
we uniformly sample values from the interval $[(1-\theta)\param, (1+\theta)\param]$,
where $\theta$ is a parameter denoting the maximum percentage change.

Figure~\ref{fig:robust} summarizes the results for the three primary models.
 We plot the number of instances that \emph{succeed} with a probability $\geq 0.8$
(i.e., are still solutions more than $80\%$ of the time)
as the amount of noise $\theta$ increases.
Obviously, when $\theta$ is 0, all instances succeed with probability 1;
as $\theta$ increases, the success rate of a number of instances falls below $0.8$.
We notice that the German and Fannie Mae solutions are quite robust,
while only half of the Adult solutions can tolerate $\theta=0.03$ noise.

While our problem formulation does not enforce robustness of solutions,
the results show that most instances are quite robust to noise.
We attribute this phenomenon to the Carlini--Wagner relaxation.
Recall that we minimize $\max(0,\smodel(\inst_k)_0 - \smodel(\inst_k)_1)$.
So solutions need only get to a point where $\smodel(\inst_k)_0 \leq \smodel(\inst_k)_1$---intuitively,
barely beyond the decision boundary.
However, we notice that, for most instances, solutions end up far from the boundary.
Specifically, for the two models with more robust solutions, German and Fannie Mae,
the average relative difference between $\smodel_1$ and $\smodel_0$---i.e., $(\smodel(\inst_k)_1 - \smodel(\inst_k)_0) / \smodel(\inst_k)_1$---is 3.41 (sd=6.79) and 1.92 (sd=5.40), respectively. For Adult, the average relative difference is much smaller, 0.09 (sd=0.06), indicating that most solutions where quite close to the decision boundary, explaining their sensitivity to noise.

It would be interesting to further improve robustness by incorporating it as a first-class constraint in our problem, e.g., by reformulating Problem~\ref{eq:cw} as a \emph{robust optimization}~\cite{ben2009robust}
problem, so as we only discover solutions that are tolerant to noise.
We plan to investigate this in future work. 

\paragraph{Further Demonstration and Discussion}
To further explore applications of our algorithm, 
we consider the Drawing Recognition model~\cite{zhang2018interpreting}.
Each drawing in this model is composed of up to 128 straight line strokes.
We considered 16 sketches of cats that are not classified as cats by this model.
We constructed an action that adds a single line stroke,
where the parameters to the action are the source and target $(x,y)$ coordinates.
To ensure that the results of the actions are visible to the human eye, we add the precondition
that stroke length is between 0.1 and 0.6 in length, where the image is $1\times 1$.
The cost of an action is the length of the stroke.

We ran our algorithm up to and including length 6 ($\score$ has no effect since there is a single action). Our algorithm managed to synthesize action sequences for 11/16 instances. Three representative solutions are shown in Figure~\ref{fig:sketch}.
The first solution, of length 1, appears to be an additional whisker to the cat.
The second and third solutions, lengths 4 and 6, appear to be more arbitrary and thus may be more adversarial in nature.

Note that our task is qualitatively different from \cite{zhang2018interpreting}.
They want to find the closest image across the decision boundary
that has an $\epsilon$-ball around it.
So they start with an adversarial example and incrementally expand it into a \emph{region} of examples.
Our problem is motivated by application of real-world actions, and therefore we search for a sequence of actions to modify an image.

The results of our technique on the Drawing Recognition model may seem to suggest that the solutions are adversarial in nature. However, it is hard to formally characterize the difference between an adversarial attack and a reasonable action sequence. In image-based attacks, it’s easy to tell if the modification is meaningful, but generally, e.g., in loans, this can probably be addressed by a domain-expert on a case-by-case basis. Observationally, we see that all our results look reasonable, i.e., there are no actions of the form \textit{modify X by $\epsilon$}, where $\epsilon$ is very small for practical purposes. Moreover, our experiments show that most synthesized sequences are robust to random perturbations in their parameters, suggesting that they are not adversarial corner cases. 
One concrete way we can protect against generating adversarial feedback is to restrict our technique to models that are trained to be \emph{robust} against adversarial attacks~\cite{madry2018towards}---however, the definition of robustness will have to be tailored to the specific domain. 

The seemingly unrealistic strokes produced as solutions in the Drawing Recognition model may stem from the fact that the cost function simply penalizes the length of the stroke and in no way drives the drawing of a `realistic' stroke (In fact, it is not obvious how one can specify a cost function that encourages strokes which look realistic). The demonstration of our technique on this dataset serves to exhibit the versatility of our problem setting and proposed solution. 
\section{Related Work}\label{sec:relatedwork}

We focus on  works  not discussed in Section~\ref{sec:introduction}.

\paragraph{Interpretable Machine Learning}
Recently, there has been a huge interest in explainability in machine learning,
particularly for deep neural networks.
Most of the works have to do with \emph{highlighting} the important features that led to a prediction,
e.g., pixels of an image or words of a sentence.
For instance, the seminal work on LIME~\cite{ribeiro2016should} trains a simple local classifier
and uses it to rank features by importance---many other works employ different techniques to 
hone in on important features, e.g., ~\cite{datta2016algorithmic,sundararajan2017axiomatic,lundberg2017unified}.
This is usually not enough: knowing, for instance, that your credit score affected the loan decision does not tell you how much you need to increase it by to be eligible for a loan, or whether there are other actions you can take.
This is the distinguishing aspect of our work---providing actionable feedback.

\paragraph{Program Synthesis}
We view our algorithm through the lens of program synthesis.
Our algorithm is a form of enumerative program synthesis, a simple paradigm that has shown to be performant in many domains---see \cite{alur2018search} for an overview.
Our work is also related to differentiable programming languages~\cite{reed2015neural,bovsnjak2017programming,pmlr-v70-gaunt17a,gaunt2016terpret,valkov2018houdini}.
The idea  is to use numerical optimization to fill in the holes (parameters) in differentiable programs.
Our work is similar in that we define a differentiable language of actions and costs and use
numerical optimization to learn appropriate parameters for those actions.
At an abstract level, our algorithm is similar in nature to that of \cite{valkov2018houdini}.
They enumerate functional programs over neural networks and use optimization to learn parameters;
here, we enumerate action sequences and use optimization to learn their parameters.
There is also a growing body of work on using deep learning to perform and guide program synthesis, e.g., ~\cite{parisotto2016neuro,chen2018execution,bunel2018leveraging}

Symbolic synthesis techniques typically use SAT/SMT solvers to search the space of programs~\cite{solar2006combinatorial,gulwani2011synthesis}.
Unfortunately, the range of programs they can encode is limited by decidable and practically efficient theories.
While our problem can be encoded as an optimal SMT problem in linear arithmetic~\cite{li2014symbolic}---equivalently, an MILP problem---we will have to restrict all actions, costs, and models to be linear.
In practice, this is quite restrictive. While our approach is incomplete, unlike in decidable first-order theories, it offers the flexibility and generality of being able to handle arbitrary differentiable models and actions.

\paragraph{Planning and Reinforcement Learning}
Our problem can also be viewed as a planning problem in a continuous (or hybrid) domain.
Most such planners are restricted to linear domains that are favorable to an SMT or MILP encoding
or restricted forms of non-linearity, e.g.,~\cite{cashmore2016compilation,piotrowski2016heuristic,bryce2015smt}.
Some such planners also combine search and a form of optimization, typically LP, e.g., \cite{fernandez2018scottyactivity,coles2012colin}.
Recently, \cite{wu2017scalable} presented a scalable approach by reducing the planning problem to optimization and solving it with TensorFlow.
In their setting, they deal with simple domains, e.g., 1 action; they do not have a goal state (in our case changing classification of a neural network), just a reward maximization objective; and they do not incorporate preconditions.
Further, they are interested in very long plans over some time horizon, while our focus on generating small, actionable plans.

Our problem is also related to reinforcement learning (RL) with continuous action spaces, e.g., \cite{lillicrap2015continuous}.
The power of RL is its ability to construct a general policy that can lead to a goal state.
Thus, given a model, it would be interesting to use RL to learn a single policy
that we can then apply to any input to the model, in contrast with our approach that learns a specific sequence of actions for every input.
%


\section{Conclusion}\label{sec:conclusion}
We described a solution to the problem of presenting simple actionable feedback to subjects of a decision-making model so as to favorably change their classification.
We presented a general solution where a domain expert specifies a differentiable set of actions that can be performed along with their costs.
Then, we combine a search-based technique with an optimization technique to construct an optimal sequence of actions that leads to classification change.
Our results demonstrate the promise of our technique and its applicability.
There are many potential avenues for future work, e.g., exploring effects of different relaxations on optimality and robustness of results and adapting the algorithm to a complete black-box setting where we can only query the model.
Another interesting avenue is to explore how our approach can be used to \emph{game} the system in a \emph{strategic classification} setting~\cite{Hardt:2016:SC:2840728.2840730}.

\bibliographystyle{unsrt}
\bibliography{main}

\end{document}


\maketitle

\appendix

\section{Further implementation details}

\subsection{Actions, Costs, and Preconditions}
Our algorithmic framework is built for complex actions, on which logical constraints can be imposed along with custom cost functions, some details of which are described below:  
\begin{itemize}

    \item Continuous Actions: These are actions which modify continuous (numeric) input features. 
    \begin{itemize}
        \item Continuous actions are parameterized by one or more real numbers. The values of the parameters directly modify the feature values of the related features. For example, the parameter corresponding to `Change Working Hours' action in the Adult Income model directly adds to the original feature value.  
        \item Note that numeric features are normalized before being fed into neural network models. Our framework is built for easily defining cost functions as well as conditions for the actions in either feature space (normalized or unnormalized). 
        \item A continuous action can also modify more than one input feature. For example, the `Adjust Loan Period' action in the German Loan model modifies both the `Loan Duration' and `Credit Amount' input features, while maintaining the ratio between them.
    \end{itemize}
    \item Categorical Actions: Categorical actions modify categorical input features, which are either binary or one-hot encoded. Such actions are non-differentiable, and therefore have no parameters that require optimization. These actions result in a deterministic change in the output features, to the desired binary or one-hot encoding. An example of a categorical action is the `Get Guarantor' in the German Loan model.

    \item Cost functions: Our framework allows the user to define arbitrary cost functions for each action, which reflect the relative difficulty of performing different actions. Examples of cost functions we have used in our experiments include the $L_1$ loss, $L_2$ loss and constant loss.
\end{itemize}

\subsection{Hyperparameters / Optimization}
\paragraph{Preconditions} When relaxing a precondition of the form $x > c$, we introduce two
hyperparameters $\tau$ and $\tau'$, as discussed in text: $\pre'_i(\inst,\param) = \tau\exp(-\tau'(x-c))$. 
The hyperparameters $\tau$ and $\tau'$ determine the steepness of the continuous boundaries. $\tau$ determines the cost value at the specified boundary $c$, whereas $\tau'$ determines the slope of the boundary function. Depending on the size $s = x_{max} - x_{min}$ of the feature domain on which the conditions are applied, we scale the two hyperparameters such that the boundary effect begins at approximately $c + 0.01s$. Specifically, we use the following formulation:
    $\tau = \tau' = 1000/{s}$.

\paragraph{Carlini--Wagner}
The Carlini--Wagner relaxation introduces hyperparameter $c$. We modify $c$ at variable intervals $t$ during the search. Specifically, in practice, we perform the following operation at each interval:
\begin{enumerate}
  \item if boundary is reached $\rightarrow c = c / 10;\ t = t * 2$
  \item if in 0 classification region and never reached the boundary $\rightarrow c = c * 2$
  \item if in 0 classification region and previously reached the boundary $\rightarrow c = c * 2;\ t = t / 2$
\end{enumerate}
Initially, $c = 1e5$ and $t=100$, and we maintain that  $1e-5 \leq c \leq 1e10$.

\paragraph{Optimization iterations}
Adam optimizer is run for maximum of \num{10000} iterations. Every 100 iterations it is checked if the adversarial cost term (i.e. $ h(x_k) $ as defined in Section 3.1) has decreased by at least $10^{-4}$, otherwise we terminate it.

\newpage
\section{Dataset and Action descriptions}

Preconditions are split into Precondition1 $\wedge$ Precondition2.
Precondition1 denotes 
constraints that refer to the properties of the original features. For example, the precondition for the `Get Guarantor' action is the absence of a guarantor in the first place. 
Precondition2 refers to the properties of the transformed features after the action is performed. For example, the modified credit score after the `Change Credit Score' action in the Mortgage Underwriting model should be between 300 and 850. 
     
\begin{table}[h]
    \centering
    \begin{tabular}{ p{1.9cm}  p{3.3cm}  p{1.8cm}  p{1.95cm}  p{3cm}}
        \toprule
        Model & Action & Action type & Precondition1 & Precondition2 \\
        \midrule
        German & Change Credit Amount & Continuous & Age>15 & 0<Cr.Amt<100K\\
        & Change Loan Period & Continuous & - & 0<Loan.Pr.<120\\
        & Adjust Loan Period & Continuous & Cr.Amt>1000 & 0<Cr.Amt<100K $\wedge$ 0<Loan.Pr<120\\
        & Wait Years & Continuous & - & Cur.Age<Age<120\\
        & Naturalize & Categorical & Not a citizen & -\\
        & Get Unskilled Job & Categorical & Unemployed & - \\
        & Get Guarantor & Categorical & No current guarantor & -\\
        \hline
        Adult  & Wait Years & Continuous & - & Cur.Age<Age<120\\
        & Change Working Hours & Continuous & - & 0<Work.Hrs<90\\
        & Increase Capital Gain & Continuous & Cap.Loss<1 & 0<Cap.Gain<100000\\
        & Change Capital Loss & Continuous & Cap.Gain<1 & 1<Cap.Loss<5000\\
        & Add Education & Continuous & - & Cur.Age<Age<120 $\wedge$ Cur.Edu<Edu<16.5\\
        & Enlist & Categorical & Not currently enlisted & -\\
        \hline
        Fannie Mae & Change Credit Score & Continuous & - & 300<Cr.Score<800\\
        & Change Num Units & Continuous & - & 0<Num.Units<5\\
        & Change Debt-to- \qquad Income Ratio & Continuous & - & 0<DTI<100\\
        & Change Interest Rate & Continuous & - & 0<Int.Rate<30\\
        & Change Loan Term & Continuous & - & 0<Loan.Term<800\\
        \bottomrule
    \end{tabular}
    \caption{Action descriptions for the three primary models. Actions for the Drawing Recognition task are described in main text.}
    \label{tab:my_label}
\end{table}

\newpage
\section{Full Drawing Recognition Results}
\begin{figure}
    \centering
        \includegraphics[scale=0.2]{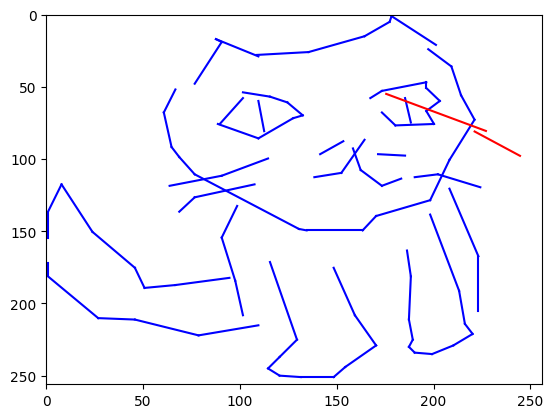}
        \includegraphics[scale=0.2]{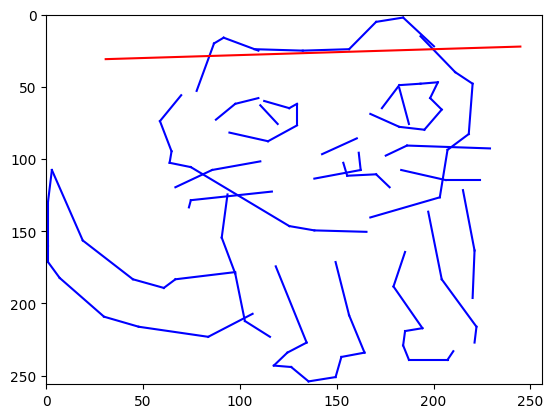}
        \includegraphics[scale=0.2]{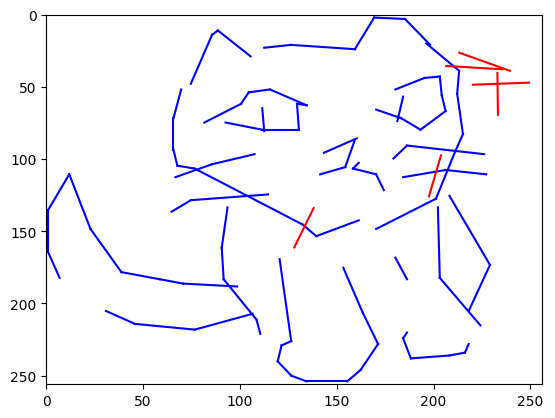}
        \includegraphics[scale=0.2]{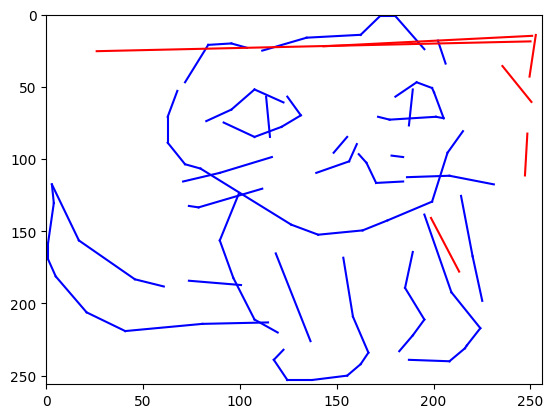}
        \includegraphics[scale=0.2]{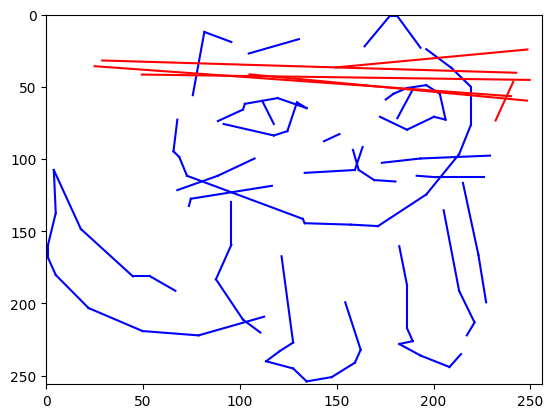}
        \includegraphics[scale=0.2]{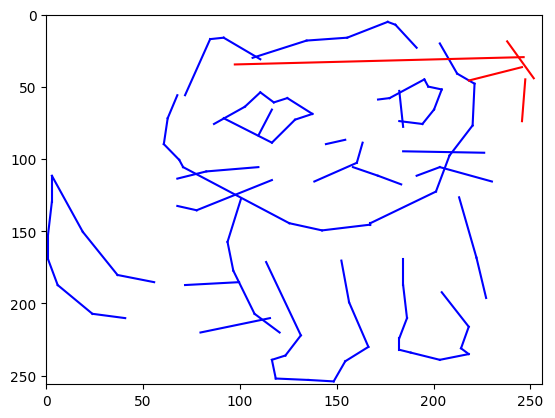}
        \includegraphics[scale=0.2]{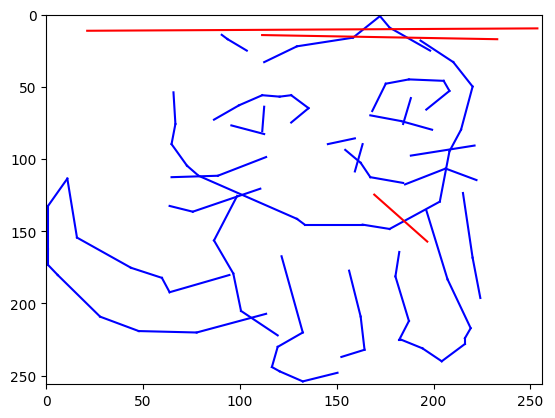}
        \includegraphics[scale=0.2]{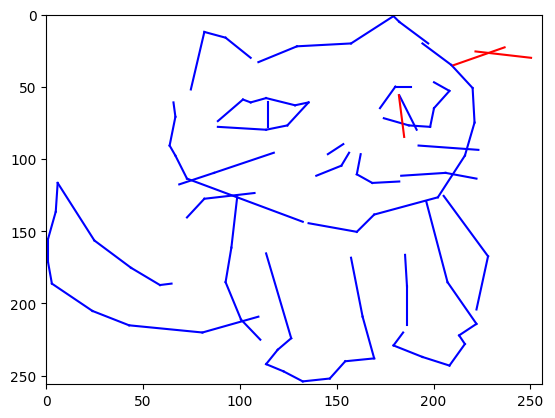}
        \includegraphics[scale=0.2]{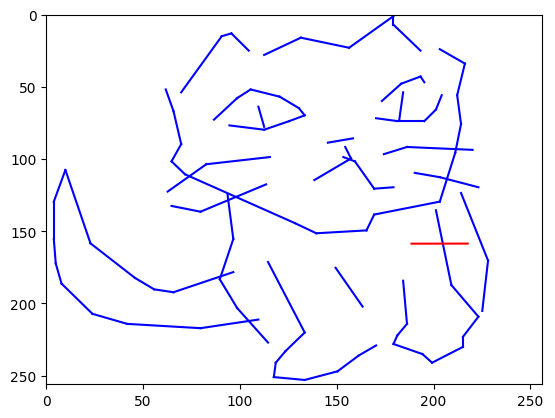}
        \includegraphics[scale=0.2]{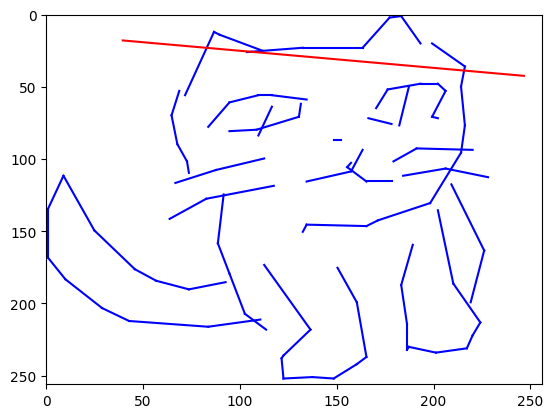}
        \includegraphics[scale=0.2]{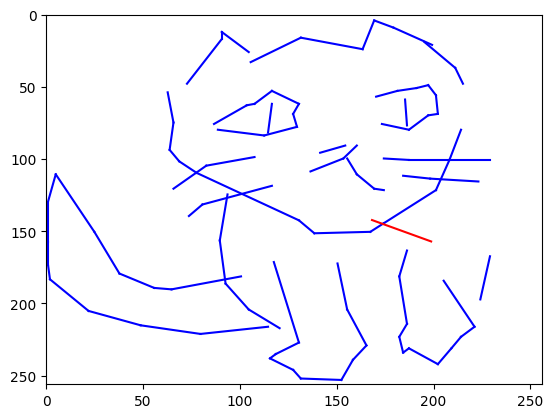}
\end{figure}